\documentclass{article}

\usepackage{microtype}
\usepackage{graphicx}
\usepackage{subfigure}
\usepackage{booktabs} 
\usepackage{amsmath} 
\usepackage[semicolon]{natbib}

\usepackage{tikz}
\usepackage{pgfplots}
\pgfplotsset{compat=1.12}
\usepgfplotslibrary{groupplots}
\usepackage{amssymb}
\usepackage{bm,amsfonts}
\usepackage{filecontents}
\graphicspath{ {./images/} }
\usepackage{enumitem}
\usepackage{breqn}
\usepackage{subfigure}
\usepackage{float}
\usepackage{multirow}
\usepackage{booktabs}

\usepackage{hyperref} 

\usepackage[accepted]{icml2019}

\definecolor{ppurple}{HTML}{9F4C7C}
\definecolor{ggreen}{HTML}{13C187}

\begin{document}

\twocolumn[
\title{Smoothed Gaussian Mixture Models for Video Classification and Recommendation}
\author{Sirjan Kafle\footnotemark, LinkedIn, skafle@linkedin.com \\ Aman Gupta\footnotemark[\value{footnote}], LinkedIn, amagupta@linkedin.com \\ Xue Xia\footnotemark[\value{footnote}], LinkedIn, xuxia@linkedin.com \\ Ananth Sankar, LinkedIn, ansankar@linkedin.com \\ Xi Chen\footnotemark, chenxi199008@gmail.com \\ Di Wen, LinkedIn, dwen@linkedin.com \\ Liang Zhang\footnotemark[\value{footnote}], Pinterest, liangzhang@pinterest.com}
\date{\vspace{-0.2in}}
\maketitle

\begin{abstract}
Cluster-and-aggregate techniques such as Vector of Locally Aggregated Descriptors (VLAD), and their end-to-end discriminatively trained equivalents like NetVLAD have recently been popular for video classification and action recognition tasks. These techniques operate by assigning video frames to clusters and then representing the video by aggregating residuals of frames with respect to the mean of each cluster. Since some clusters may see very little video-specific data, these features can be noisy. In this paper, we propose a new cluster-and-aggregate method which we call smoothed Gaussian mixture model (SGMM), and its end-to-end discriminatively trained equivalent, which we call deep smoothed Gaussian mixture model (DSGMM). SGMM represents each video by the parameters of a Gaussian mixture model (GMM) trained for that video. Low-count clusters are addressed by smoothing the video-specific estimates with a universal background model (UBM) trained on a large number of videos.  The primary benefit of SGMM over VLAD is smoothing which makes it less sensitive to small number of training samples. We show, through extensive experiments on the YouTube-8M classification task, that SGMM/DSGMM is consistently better than VLAD/NetVLAD by a small but statistically significant margin. We also show results using a dataset created at LinkedIn to predict if a member will watch an uploaded video.
\\

\end{abstract}
\vspace{0.4in}
]

\addtocounter{footnote}{1}
\footnotetext{Equal contribution}
\addtocounter{footnote}{1}
\footnotetext{Work done while at LinkedIn}
\clearpage

\section{Introduction}
Video content is increasingly becoming the {\em de facto} method for consuming information for entertainment, education, and news. Coupled with the advent of advanced computer vision algorithms~\cite{alexnet, vggnet, inceptionv3, fasterrcnn} and the availability of large-scale labeled video datasets \cite{youtube8m, UCF101, HMDB51}, this has resulted in strong research interest in areas of video understanding, such as video classification and action recognition. A crucial aspect of these problems is the learning of fixed-dimensional vector representations (or features) that capture the content in videos. Cluster-and-aggregate techniques have recently become the basis of many such representations~\cite{allaboutvlad, vlad, netvlad, actionvlad, willow}.

In this paper, we propose a new cluster-and-aggregate method for video representation, which we call Smoothed Gaussian Mixture Model (SGMM). We also introduce its discriminative, end-to-end trained counterpart, which we call Deep Smoothed Gaussian Mixture Model (DSGMM). SGMM uses the parameters of a Gaussian mixture model (GMM) to represent each video, relying on the parameters of a universal background model (UBM) when the video-specific GMM clusters receive little data. Smoothing GMM-based representations in this manner leads to more robust video representations. 

The novelty of our approach lies in the use of a principled smoothing approach to derive robust estimates of statistics in the cluster-and-aggregate framework.\footnote{In this work, we use GMMs to implement soft clustering.} Previous video understanding research using this framework has focused on computing different cluster statistics to represent the video. This includes Bag of Words (BoW)~\cite{csurka2004visual,philbin2008lost}, which computes the cluster mass or zeroth-order statistic, Vector of Locally Aggregated Descriptors (VLAD)~\cite{vlad, allaboutvlad}, which computes the first-order mean-centered statistic, Fisher Vectors (FV)~\cite{perronnin2010improving}, which uses the first and second-order mean-centered statistics, and RVLAD~\cite{willow}, which computes the uncentered first-order statistic. 
Our contribution is orthogonal to the specific statistics themselves -- instead, we use Bayesian smoothing of GMM parameters~\cite{gauvain1994maximum,reynolds2000speaker} to compute robust estimates of all these previous statistics. We evaluate our proposed SGMM-based video representations on the task of video classification using the popular YouTube-8M dataset \cite{youtube8m} and demonstrate improvements over previous cluster-and-aggregate techniques.  

The paper is organized as follows: In Section \ref{sec:related-work}, we discuss related work; in Section \ref{sec:our-approach} we describe in detail our new SGMM/DSGMM approach; in Section \ref{sec:experiments} we present comprehensive experiments on the public YouTube-8M dataset, and our internal LinkedIn dataset, and finally in Section \ref{sec:conclusion} we give our conclusions and pointers to future work.

\section{Related Work} 
\label{sec:related-work}

Most current methods compute a video-level representation by combining representations of the time-sampled video frames. The sampling rate may depend on the task; for example, for video classification one frame per second is sufficient~\cite{youtube8m}.
The video frames are processed using deep convolutional neural networks (CNN), which have demonstrated strong performance for image classification~\cite{alexnet, vggnet, inceptionv3, szegedy2017inception}.  
Video processing methods commonly use the last layer before the classification layer of these CNNs to compute $D$-dimensional frame-level features~\cite{youtube8m,wu2019long}. 

To produce video features, some methods model temporal structure in videos using recurrent neural networks (RNN). In particular, Long-Short Term Memory (LSTM)~\cite{lstm} networks have been used to process the frame-level features, with the last hidden state of the LSTM serving as the video feature~\cite{donahue2015,sun2016exploiting,srivastava2015unsupervised}. Another approach is to apply a temporal attention mechanism to select/weigh different temporal segments to generate a video level feature~\cite{yao2015describing}. Other CNN-based approaches combine spatial information~\cite{karpathylargescale,tran2018closer} or use 3D convolutions to capture the spatio-temporal information in a video~\cite{c3d,ji20123d}. 

Methods based on aggregation treat the video as a set of frames, ignoring the temporal order. The simplest of these include average or max pooling of the $D$-dimensional frame features~\cite{youtube8m} to produce a $D$-dimensional video feature. Cluster-and-aggregate methods first group the $D$-dimensional frames into $K$ different clusters, and then aggregate frame statistics for each cluster. BoW is the simplest such technique, where each video is represented by a histogram of frame counts over the clusters~\cite{youtube8m,willow,sivic2003video}, resulting in a $K$-dimensional representation. A popular recent technique, VLAD~\cite{vlad, allaboutvlad} instead computes the vector difference between each video frame and its corresponding cluster centroid and then aggregates these residuals for each centroid, resulting in a $(K \times D)$-dimensional representation.

Finally, VLAD has been integrated into an end-to-end supervised training framework in methods like NetVLAD~\cite{netvlad} and ActionVLAD~\cite{actionvlad}. The idea behind these methods is to use soft clustering, which allows us to use backpropagation to train the clusters in a supervised manner, instead of using the unsupervised $K$-means algorithm. The cluster assignment probability depends on the parameters of the clusters. However, to simplify matters, NetVLAD decouples the computation of the assignment probability from the cluster parameters. It is also possible to train the cluster parameters while maintaining the coupling~\cite{variani2015gaussian,DBLP:journals/corr/WieschollekGL17}. Features like BoW and FV have also been integrated into the NetVLAD end-to-end framework~\cite{willow}, resulting in NetBOW and NetFV.

Our work falls into the cluster-and-aggregate category. However, while previous works focused on the statistics that are aggregated~\cite{vlad, netvlad, willow, actionvlad}, we use a principled Bayesian smoothing approach~\cite{gauvain1994maximum} to compute robust estimates of these statistics. This is motivated by earlier work in speaker recognition~\cite{reynolds2000speaker} where such models were successfully used over a long period of time~\cite{kinnunen2010overview}.

\section{Methodology}
\label{sec:our-approach}
In Section~\ref{subsec:SGMM}, we describe our SGMM representation. We then show how it relates to the popular VLAD approach~\cite{vlad, allaboutvlad} in Section \ref{subsec:relation}, and extend it using end-to-end training like NetVLAD~\cite{netvlad, willow} in Section \ref{subsec:NetVLAD}. In Section~\ref{sec:socialnetwork} we describe how we use video representations for predicting whether a video will be watched at LinkedIn.

\subsection{Smoothed Gaussian mixture models (SGMM)} \label{subsec:SGMM}
The GMM, expressed by Equation~\ref{eq:gmm}, is a simple density function commonly used to approximate arbitrary data distributions.
\begin{equation}
p(\bm{x}) = \sum_{k=1}^{K} w_k N(\bm{x};\bm{\mu}_k,\bm{\Sigma}_k)
\label{eq:gmm}
\end{equation}
Each cluster, $k\in \{1,\ldots, K\}$, has a prior probability given by $w_k$. The clusters are represented by Gaussian distributions, $N(\bm{x};\bm{\mu}_k,\bm{\Sigma}_k)$, with mean and covariance given by $\bm{\mu}_k$ and $\bm{\Sigma}_k$. By increasing the number of Gaussians, $K$, we can approximate distributions of different shapes. GMMs have historically been used for classification tasks like speech and speaker recognition, as well as for clustering. Each Gaussian, $k$, models a cluster, and the data, $\bm{x}$, is assigned to it according to the following posterior probability:
\begin{equation}
P(k|\bm{x}) = \frac{w_k N(\bm{x};\bm{\mu}_k,\bm{\Sigma}_k)}{\sum_{k=1}^{K} w_k N(\bm{x};\bm{\mu}_k,\bm{\Sigma}_k)}
\label{eq:gmm-posterior-prob}
\end{equation}
$K$-means clustering is a simplification of GMMs, where all Gaussians share an identity covariance matrix, $\bm{I}$, the mixture weights, $w_k$, are uniform, and the data, $\bm{x}$, is assigned only to the cluster that has the highest posterior probability, $P(k|\bm{x})$. GMMs are usually trained with maximum likelihood (ML) estimation using the expectation-maximization (EM) algorithm~\cite{dempster1977maximum}, but they can also be trained for specific tasks using discriminative training~\cite{heigold2012discriminative,variani2015gaussian}.

In our work, we propose to represent a video by the parameters of a GMM. In particular, we assume we have a universal background model (UBM), which is a ``global" GMM trained on a large number of frames from many different videos. As shown in Equation~\ref{eq:gmm}, its parameters are $w_k$, $\bm{\mu}_k$, and $\bm{\Sigma}_k$. 
Each video, $v$, is composed of a set of sampled frames, $\{\bm{x}^v_1, \bm{x}^v_2, .., \bm{x}^v_{T_v}\}$, which are computed, for example, by using deep CNNs~\cite{inceptionv3, vggnet}.
The video frames $\bm{x}^v_t$ are used to estimate the video-specific GMM parameters,  $w_{k}^v$, $\bm{\mu}_{k}^v$, and $\bm{\Sigma}_{k}^v$ for $k=1$ to $K$. By convention, we superscript video-specific GMM parameters with $v$ to distinguish them from UBM parameters. To obtain the estimates of these parameters, we first compute the following sufficient statistics for the video $v$:
\begin{align}
n^v(k) &= \sum_{t=1}^{T_v} P(k|\bm{x}^v_t) \\
S_{x}^v(k) &= \sum_{t=1}^{T_v} P(k|\bm{x}^v_t) \bm{x}^v_t \label{eq:mean-suff-stats}\\
S_{x^2}^v(k) &= \sum_{t=1}^{T_v} P(k|\bm{x}^v_t) {\bm{x}^v_t}{\bm{x}^v_t}^T
\end{align}
where $P(k|\bm{x}^v_t)$ is given by Equation~\ref{eq:gmm-posterior-prob} and is the posterior probability of assigning a frame $x_t^v$ to the $k$'th cluster of the already trained UBM. These sufficient statistics can be directly used to compute the ML estimates of $w_{k}^v$, $\bm{\mu}_{k}^v$, and $\bm{\Sigma}_{k}^v$, as follows:
\begin{align}
w_{k}^v &= \frac{n^v(k)}{\sum_k n^v(k)} \label{eq:unsmoothed-wgt}\\
\bm{\mu}_{k}^v &= \frac{S_{x}^v(k)}{n^v(k)} \label{eq:unsmoothed-mean}\\
\bm{\Sigma}_{k}^v &= \frac{S_{x^2}^v(k)}{n^v(k)} - {\bm{\mu}_{k}^v}{\bm{\mu}_{k}^{v}}^T  \label{eq:unsmoothed-cov}
\end{align}
However, if $n^v(k) $ is small relative to other clusters, then this can lead to noisy estimates. To address this, we use Bayesian smoothing with respect to the UBM parameters to compute robust estimates for the video GMM as shown in~\cite{gauvain1994maximum, reynolds2000speaker}:
\begin{align}
w_{k}^v &= \lambda^v_{w,k} \frac{n^v(k)}{\sum_k n^v(k)} +(1-\lambda^v_{w,k}) w_k \label{eq:smoothed-wgt}\\
\bm{\mu}_{k}^v &= \lambda^v_{\bm{\mu},k} \frac{S_{x}^v(k)}{n^v(k)} +(1-\lambda^v_{\bm{\mu},k}) \bm{\mu}_k \label{eq:smoothed-mean}\\
\bm{\Sigma}_{k}^v &= \lambda^v_{\bm{\Sigma},k} \frac{S_{x^2}^v(k)}{n^v(k)} + (1- \lambda^v_{\bm{\Sigma},k}) (\bm{\mu}_k\bm{\mu}^T_k + \bm{\Sigma}_k) - {\bm{\mu}_{k}^v}{\bm{\mu}_{k}^{v}}^T  \label{eq:smoothed-cov}
\end{align}
The weighting terms, $\lambda^v_{w,k}$, $\lambda^v_{\bm{\mu},k}$, and $\lambda^v_{\bm{\Sigma},k}$, smooth the video sufficient statistics with respect to the sufficient statistics computed from the UBM parameters. We give the expression for $ \lambda^v_{\bm{\mu},k}$ below (the others are similar):
\begin{equation}
\lambda^v_{\bm{\mu},k} = \frac{n^v(k)}{n^v(k)+\gamma_\mu} \label{eq:lambda-formulation}
\end{equation}
Note that when $n^v(k)$, the number of frames assigned to cluster $k$, is large, the smoothed estimates tend toward the video-specific ML estimates. On the other hand, when $n^v(k)$ is small, we rely more on the UBM parameters. The non-negative $\gamma_\mu$ hyperparameter controls the level of smoothing. Often the same value is used for all statistics, so we replace $\gamma_\mu$ by $\gamma$. The smaller the value of $\gamma$, the more we believe the video frames, and the larger the value, the more we believe the UBM prior. For example, when $\gamma=0$, Equations~\ref{eq:smoothed-wgt}, ~\ref{eq:smoothed-mean}, and ~\ref{eq:smoothed-cov} reduce to the video-specific ML estimates, and when $\gamma=\infty$, they reduce to the parameters of the UBM.

In this paper, we use \textbf{only the mean parameters} $\bm{\mu}_{k}^v$ in (Equation~\ref{eq:smoothed-mean}) to represent video, $v$. This results in a $(K\times D)$-dimensional representation, where $K$ is the number of Gaussians, and $D$ is the dimensionality of the frame-level feature vector. We chose to use only the mean parameter to represent the video following ~\cite{reynolds2000speaker}, where all combinations of mixture weight, means and variances were tried and the mean was found to provide the best performance.
~
\subsection{Relation to VLAD} \label{subsec:relation}
In VLAD, the frames from each video are assigned to the UBM clusters, followed by computing the mean residuals for each cluster as follows:
\begin{equation}
VLAD_k^v = \sum_{t=1}^{T_v} P(k|\bm{x}^v_t) (\bm{x}^v_t - \bm{\mu}_k)
= S^v_x(k) - n^v(k)\bm{\mu}_k
\label{eq:vlad-2}
\end{equation}

Both VLAD (Equation~\ref{eq:vlad-2}) and SGMM (Equation~\ref{eq:smoothed-mean}) are $(K \times D)$-dimensional feature vectors, and both involve the computation of the same sufficient statistics, $S^v_x(k)$ and $n^v(k)$. However, it is likely that some Gaussians will have small counts, $n^v(k)$. VLAD simply ignores this and computes the residual feature as is; however, SGMM uses Bayesian smoothing to move the estimate in the direction of the UBM priors. In fact, a common technique done in VLAD called intra-normalization~\cite{allaboutvlad}, which takes the L2-norm of each $D$ dimensional cluster representation, exacerbates this problem by equalizing the importance of all residuals even when the corresponding counts, $n^v(k)$, are small. Also, SGMM incurs no additional complexity over VLAD, since both require the computation of exactly the same sufficient statistics, $S_x^v(k)$ and $n^v(k)$. SGMM uses an additional hyperparameter $\gamma$, but this adds no additional complexity.

\subsection{End-to-end training} \label{subsec:NetVLAD}
NetVLAD~\cite{netvlad} and ActionVLAD~\cite{actionvlad} have trained the VLAD cluster parameters along with the video classification layer in an end-to-end manner. The key to achieving this is to convert the hard assignment of data to clusters as in $K$-means, to soft assignment, which enables the computation of derivatives, and hence end-to-end training. The soft assignment of data to clusters can be done using Equation~\ref{eq:gmm-posterior-prob}. We call this the {\em coupled} approach because the cluster posterior probability is computed using the GMM model parameters. However, NetVLAD does this data assignment using the following equation:
\begin{equation}
P(k|\bm{x}^v_t) = \frac{e^{(\bm{u}_k^T \bm{x}^v_t + b_k)}}{\sum_k e^{(\bm{u}_k^T \bm{x}^v_t + b_k)}}
\label{eq:decoupled-pki}
\end{equation}
where the parameters $\bm{u}_k$ and $b_k$ are completely separate from the GMM parameters. We call this the {\em decoupled} method. While this is an over-simplification of the GMM model, it appears to work well. We note that previous work has used the {\em coupled} approach to train GMMs in an end-to-end fashion~\cite{variani2015gaussian,DBLP:journals/corr/WieschollekGL17}. 

Using exactly the same principles as NetVLAD, we extend SGMM to an end-to-end training method that we call DSGMM. We compare both the {\em coupled} and {\em decoupled} techniques in our experiments.

For the {\em coupled} approach, we explore five methods that differ in the way we constrain the GMM parameters:
\begin{enumerate}
\item
\texttt{UniformPriors}: $\forall k$, let $w_k = \frac{1}{K}$ so that they are not trainable parameters. Also, the covariance matrix is spherical and shared across Gaussians: $\forall k|\bm{\Sigma}_k = \sigma^2\bm{I}$. $\bm{\mu}_k$ and $\sigma$ are trainable parameters.

\item
\texttt{SharedSpherical}: Covariance matrix is spherical and shared across Gaussians. $w_k, \bm{\mu}_k, \sigma$ are trainable parameters.

\item
\texttt{Spherical}: Covariance matrices are spherical but not shared across Gaussians. Thus, each covariance matrix can be written as $\bm{\Sigma}_k = \sigma_k^2\bm{I}$. $w_k, \bm{\mu}_k, \sigma_k$ are trainable parameters.

\item
\texttt{SharedDiagonal}: Covariance matrix is diagonal and shared across Gaussians. $\forall k|\bm{\Sigma}_k = \text{diag}(\sigma_1^2, \sigma_2^2, \ldots, \sigma_D^2)$. $w_k, \bm{\mu}_k, \sigma_i$ for $i = 1,\ldots,D$ are trainable parameters.

\item
\texttt{Diagonal}: Covariance matrices are diagonal but not shared across Gaussians. Thus, each covariance matrix can be written as $\bm{\Sigma}_k = \text{diag}(\sigma_{k1}^2, \sigma_{k2}^2, \ldots, \sigma_{kd}^2)$. $w_k, \bm{\mu}_k, \sigma_{ki}$ for $i=1,\ldots,D$ are trainable parameters.
\end{enumerate}
The mixture weights $w_k$ must form a valid probability distribution. Furthermore, the covariance matrices for the GMMs must be positive definite. However, naive backpropagation will not ensure these constraints. We handle this using a transformation as proposed by~\cite{variani2015gaussian}. For example, in the diagonal covariance case, we maintain the following trainable parameters that are updated: $\tilde{w}_k, \tilde{\sigma}_{ki}$, and undergo the transformations $w_k = e^{\tilde{w}_k}/\left(\sum_{j=1}^Ke^{\tilde{w}_j}\right)$ and $\sigma_{ki} = e^{\tilde{\sigma}_{ki}}$.
During optimization, we backpropagate gradients all the way through these transformations. 

Along with the {\em decoupled} approach of NetVLAD, the above five {\em coupled approaches} can also be used to compute NetVLAD and DSGMM features within the end-to-end framework. We compare all these approaches in our experiments (see Section~\ref{sec:experiments}).

\begin{figure*}
\centering
\includegraphics[width=0.8\textwidth]{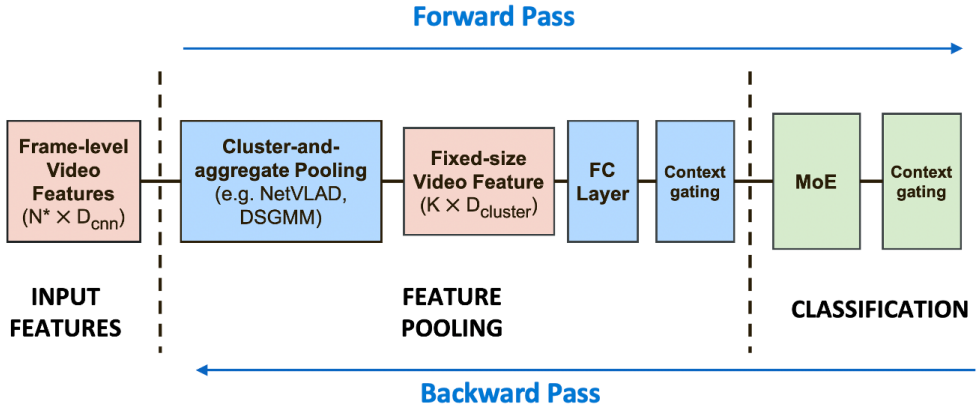}
\caption{End-to-end architecture for the network that is trained using cross-entropy loss per label for video classification. Frame level features are used in the cluster-and-aggregate pooling layer which outputs a fixed dimensional video representation. This representation is passed through context gating and mixture of experts layers. Blue and green components indicate layers with trainable parameters. Diagram modified from~\cite{willow}}\label{fig:e2etraining}
\end{figure*}

These techniques can be thought of as a \textit{cluster-and-aggregate pooling} layer in the full video classification network architecture which is summarized in Figure~\ref{fig:e2etraining}. The output from the \textit{cluster-and-aggregate pooling} layer is a $K\times D$ representation for each video. This is used in successive context gating and mixture of experts (MoE) following the technique in~\cite{willow}.

\subsection{Video recommendation} \label{sec:socialnetwork}
\subsubsection{ Co-watch-based embeddings}\label{sec:cowatch-embeddings}
Consider the task of personalized video recommendation on the LinkedIn feed, based on a user's past video watch behavior. A video watched by a user $U$ for more than a threshold $T$ seconds is labeled a \textit{valid video watch}. Given the set $S_U$ of valid video watches for $U$, we are interested in predicting whether the user will watch a candidate video $v_c$ or not. We start by defining co-watched videos as a pair of videos that have a valid watch by the same member in a short period of time (we use 30 minutes). Assume our base video feature vector computed using NetVLAD or DSGMM is given by $\bm{g}(v)$. We transform this feature into an embedding space using a transform $\bm{f}(\bm{g}(v))$, where videos that are co-watched are closer together, and those that are not are farther apart. We ensure that $\bm{f}(\cdot)$ is L2-normalized. We then use these video embeddings as features to predict if a member will watch the given candidate video. 

To learn the transformation $\bm{f}(\cdot)$ we take inspiration from a recent work \cite{cdml}, where the authors use a triplet-loss-based formulation~\cite{tripletloss} to embed videos into a co-watched embedding space. Specifically, we optimize the following triplet-loss based function:
\begin{multline} \label{eq:tripletLoss}
\mathcal{L}(f) = \sum_{i=1}^{N} \max(\|\bm{f}(\bm{g}(v^i_a)) - \bm{f}(\bm{g}(v^i_p))\|^2  -  \\ \|\bm{f}(\bm{g}(v^i_a)) - \bm{f}(\bm{g}(v^i_n))\|^2 + \alpha , 0)
\end{multline}
where $\bm{g}(v^i_a)$ is the \textit{anchor} video feature vector in the $i^{th}$ triplet. $\bm{g}(v^i_p)$ is the feature vector for a \textit{positive} video frequently co-watched with the anchor. $\bm{g}(v^i_n)$ is the feature vector of a \textit{negative} video which is not co-watched with the anchor despite being presented to the user within a 30 minute window of the anchor. $||\cdot||$ represents the Euclidean distance, and $\alpha$ is a margin hyperparameter. For $\bm{f}(\cdot)$ we use a two-layer feedforward neural network which is trained with $\bm{g}(\cdot)$ in an end-to-end fashion. Thus, we effectively train the cluster-and-aggregate pooling layer to create an embedding catered for video recommendation.

\subsubsection{Predicting video watch by user} \label{sec:videowatchpredict}
Using the video embedding features $\bm{f}(\cdot)$ from the previous section, we now describe two methods to predict whether a user will watch a given candidate video $v_c$. In the first approach, let $v_h$ be a video in user $U$'s watch history set $S_U$. We compute cosine-similarity scores for all $v_c, v_h$ pairs and compute two aggregations of these scores:
\begin{equation} \label{eq:avgsim}
Avg(U,v_c) = \frac{1}{|S_U|}\sum_{v_h \in S_U} cos(\bm{f}(\bm{g}(v_c)),  \bm{f}(\bm{g}(v_h)))
\end{equation}
\begin{equation} \label{eq:maxsim}
Max(U,v_c) =  \max_{v_h \in S_U} cos(\bm{f}(\bm{g}(v_c)),  \bm{f}(\bm{g}(v_h)))
\end{equation}

In our second approach, we use the generalized linear mixed (GLMix) model~\cite{zhang2016glmix}. This uses the following logistic regression model:
\begin{equation}
\mathrm{logit}(p_U(v_c)) = \beta_0 + \beta_{U,0} + {\bm{f}(\bm{g}(v_c))}^T\bm{\beta}_U
\end{equation}
where $p_U(v_c)$ is the probability that user $U$ watches video $v_c$; the logit function $\mathrm{logit}(p) = \log \frac{p}{1-p}$ is the link function; $\beta_0$ indicates the global intercept; $\beta_{U,0}$ indicates user $U$'s propensity to watch any video, and $\bm{\beta}_U$ represents the strength of user $U$'s interaction with the video embedding features $\bm{f}(\bm{g}(v_c))$.

\section{Experiments} \label{sec:experiments}
In this section, we evaluate NetVLAD and DSGMM-based architectures using a two-pronged approach. We first report results for video classification on the YouTube-8M dataset \cite{youtube8m} across a variety of models and parameter settings.  We also compare different video representations on the task of personalized video recommendations for LinkedIn users.
\subsection{Video Classification on YouTube-8M}

\subsubsection{YouTube-8M dataset}

The 2018 version of the YouTube-8M dataset ~\cite{youtube8m} was released as part of the 2nd YouTube-8M video understanding challenge ~\cite{2ndyoutubechallenge}. Videos are assigned one or more labels out of a total of 3862 labels, with an average of 3 labels per video. Examples of labels include \textit{sports}, \textit{cooking}, \textit{fashion} and \textit{electric guitar}. There are about 6.1 million videos in the dataset, ranging from 120 to 500 seconds in length. The dataset contains pre-extracted video and audio features, with video features sampled at 1 frame per second. In this paper, we focus only on frame-level video features for all experiments.

The original YouTube-8M test set does not have publicly available labels. We, therefore, use the original validation set to carve out two sizable, disjoint sets of samples for the purposes of validation and testing. Details about the dataset splits we use can be found in Table \ref{table:ourDataset}. 

\begin{table}[H]
    \begin{center}
    \begin{tabular}{ |p{2cm}|p{3cm}|  }
        \hline
        \bf{Dataset type} & \bf{No. of samples}\\
        \hline
        Training set   & $3,887,892$\\
        \hline
        Validation set & $105,147$ \\
        \hline
        Test set & $113,331$\\
        \hline
    \end{tabular}
    \end{center}
    \caption{Our splits of the YouTube-8M dataset.}
    \label{table:ourDataset}
\end{table}

\subsubsection{Hyperparameters and implementation details}

We build off the codebase from~\cite{willow}, which includes the NetVLAD component, context gating to capture dependencies among features, and a Mixture-of-Experts (MoE) as the classification component. To implement our SGMM/DSGMM component, we only change the NetVLAD code.

All the models were trained using the Adam optimizer, with learning rate set to $0.0002$ and gradients clipped to a range of $-1$ to $1$. The learning rate was decayed exponentially every $4$ million steps by a factor of $0.8$. For model selection, we chose model snapshots that yielded the lowest loss on the entire validation set. We report test metrics using the selected checkpoints. The majority of our analysis was performed with the number of clusters set to $256$, although we also study the effect of varying the number of clusters on model performance. For all architectures, we sampled $30$ frames with replacement from each video for training, validation and testing. All experiments were run on machines with one or more NVIDIA Tesla V100 GPU(s).  

\subsubsection{Evaluation metrics}
We report results using the following metrics introduced by~\cite{youtube8m}:

\begin{itemize}
    \item \textbf{Global Average Precision (GAP)} - This metric can be defined as 
    \begin{equation}
     GAP = \sum_{k=1}^{N} p(k) \Delta r(k)
    \end{equation}
     where $p(k)$ is the precision for prediction $k$ in the ranked list of predictions. $\Delta r(k)$ is the changes in recall when moving from prediction $k-1$ to prediction $k$ in the ranked list. As per the 2nd YouTube-8M challenge ~\cite{2ndyoutubechallenge}, we produce $20$ labels per video and  thus set $N$ to $20$. 
     \item \textbf{Hit@1} - This metric is the proportion of samples that contains one of the ground truth labels in the top position of the ranked list of predictions. 
\end{itemize}

\subsubsection{Model evaluation across different architectures}

We report classification performance on the YouTube-8M dataset \cite{youtube8m} using the best parameter settings for different models in Table \ref{table:mainResults}. We evaluate the following categories of models:

\begin{enumerate}[label=(\alph*)]
  \itemsep0em 
  \item Average pooling of all video frames
  \item VLAD and NetVLAD
  \item SGMM and DSGMM
\end{enumerate}

VLAD and SGMM use soft assignment to clusters as in Equation~\ref{eq:gmm-posterior-prob}. We train a GMM with $K=256$ clusters, and a full $D \times D$ covariance matrix shared across clusters ($D=1024$). If the pretrained GMM is parameterized by mixture priors $w_k$, means $\bm{\mu}_k$, and single covariance matrix $\bm{\Sigma}$, then $\bm{u}_k$ and $b_k$ are set as follows (more general formulation of the reduction found in~\cite{netvlad}):
\begin{align}
\bm{u}_k &= \bm{\Sigma}^{-1}\bm{\mu}_k \label{eq:vlad-params1} \\
b_k &= \ln w_k - \frac{1}{2}\bm{\mu}_k^T\bm{\Sigma}^{-1}\bm{\mu}_k \label{eq:vlad-params2}
\end{align}

\textbf{Main results:} As seen in Table~\ref{table:mainResults}, the simplest architecture, average pooling of frame embeddings, yields the poorest performance while DSGMM outperforms all methods for both GAP and Hit@1. The difference for DSGMM Hit@1 performance over NetVLAD on
the test set is statistically significant using McNemar’s test with $p < 0.001$. We use McNemar's test only for Hit@1 since it cannot be applied to GAP. The best performing NetVLAD and DSGMM architectures use $256$ clusters and utilize \textit{intra-norm} without a \textit{final-norm}. Furthermore, for DSGMM, we found that setting $\gamma$ to $0.125$ yielded the best performance. These settings were found through a grid-search.
\textit{Intra-norm} L2-normalizes the representation within each cluster~\cite{allaboutvlad}, whereas \textit{final-norm} applies L2-normalization to the entire representation. 
We make the following further observations from Table~\ref{table:mainResults}:

\textbf{SGMM vs. VLAD:} SGMM yields similar performance to VLAD across both GAP and Hit@1. The results for these are better than average pooling.

\textbf{End-to-end training:} DSGMM and NetVLAD outperform their unsupervised counterparts, SGMM and VLAD, showing that end-to-end training plays an important role in learning rich GMM-based video representations. Furthermore, the best setting for DSGMM outperforms the best setting for NetVLAD.

\textbf{DSGMM vs. NetVLAD:}
Table~\ref{table:mainResults} shows that DSGMM has a 0.4\% absolute improvement in GAP over NetVLAD on the YouTube-8M dataset. To put this in perspective, we compare our gains with those reported in~\cite{willow} for NetVLAD over previous methods. In our work, we used the May 2018 version of the YouTube8M dataset, whereas~\cite{willow} used the currently deprecated Feb 2017 version; therefore, we could not exactly replicate their results. However, we reran NetVLAD, NetFV, and NetBOW using the code from~\cite{willow} on our test set, and got almost identical results (Table~\ref{table:comparisonSOTA}). Thus, we believe that we can fairly compare the gains from our paper to those reported in~\cite{willow}. As summarized in Table~\ref{table:comparisonSOTA}, our 0.4\% absolute gain over NetVLAD is greater than the gains achieved by NetVLAD over both NetFV and NetRVLAD, methods of comparable complexity. When compared to NetBOW, both NetVLAD and DSGMM are significantly better, with DSGMM showing a larger gain.

\begin{table}
\begin{tabular}{c@{\quad}cc@{\quad}cc}
  \toprule
  \multirow{2}{*}{\raisebox{-\heavyrulewidth}{Model}} & \multicolumn{2}{c}{GAP} & \multicolumn{2}{c}{Hit@1} \\
  \cmidrule{2-5}
  & Validation & Test &Validation  & Test \\
  \midrule
  AvgPooling & 0.809 & 0.810 & 0.843 & 0.845 \\
    \hline
  VLAD & 0.818 & 0.819 & 0.851 & 0.852 \\
  SGMM & 0.818 & 0.819 & 0.851 & 0.853 \\
  \hline
  NetVLAD & 0.830 & 0.831 & 0.860 & 0.863 \\
  DSGMM & \textbf{0.834} & \textbf{0.835} & \textbf{0.864} & \textbf{0.866} \\
  \bottomrule
\end{tabular}
\caption{Model evaluation on the validation and test sets.}
\label{table:mainResults}
\end{table}

\begin{table}
\begin{tabular}{ c c c }
    \toprule
    \bf{Method} & \bf{Source} & \bf{Reported GAP} \\
    \hline
    NetBOW & \cite{willow} & 0.820 \\
    NetFV & \cite{willow} & 0.830 \\
    NetRVLAD & \cite{willow} & 0.831 \\
    NetVLAD & \cite{willow} & 0.832 \\
    \hline
    NetBOW & our paper & 0.820 \\
    NetFV & our paper & 0.831 \\
    NetVLAD & our paper & 0.831 \\
    DSGMM & our paper & 0.835 \\
    \bottomrule
\end{tabular}
\caption{Comparison of DSGMM gains to those reported in~\cite{willow} for NetVLAD}
\label{table:comparisonSOTA}
\end{table}

\subsubsection{Effect of varying number of clusters and normalization} \label{varykandnorm}

\textbf{Effect of number of clusters}. We experiment with varying the number of clusters for NetVLAD and DSGMM. Since the output representation of these methods are fairly large, it is important to understand the relationship between representation capacity and model performance on downstream tasks. 

\begin{figure*}[h]
    \begin{center}
    \begin{tikzpicture}[scale=0.75][font=\large]
    \begin{axis}[
        legend style={
                at={(0.9,0.3),anchor=north}
            },
       xlabel={Number of clusters $k$ },
        ylabel={Global Average Precision (GAP)},
         ymajorgrids = true,
        symbolic x coords={1, 2, 32, 64, 128, 256, 384, 511},
        xtick=data]
        nodes near coords, 
point meta=y,
        \addplot[mark=diamond*,thick,red] coordinates {
           (1,0.8037)    (2, 0.8066)    (32, 0.8212)    (64, 0.8235)    (128, 0.8270)   (256, 0.8269) (384, 0.8273) (511, 0.8287)};
        \addlegendentry{NetVLAD}
        \addplot[mark=o,mark options={solid},blue,thick,dashed] coordinates {
           (1,0.8040)    (2, 0.8093)    (32,0.8266)    (64,0.8302)    (128, 0.8319)   (256, 0.8332) (384, 0.8311) (511, 0.8308)};
        \addlegendentry{DSGMM}

    \end{axis}
    \end{tikzpicture}
    \qquad
    \begin{tikzpicture}[scale=0.75][font=\large]
    
    \begin{axis}[
        legend style={
                at={(0.9,0.3),anchor=north}
            },
       xlabel={Number of clusters $k$ },
        ylabel={Hit@1},
         ymajorgrids = true,
        symbolic x coords={1, 2, 32, 64, 128, 256, 384, 511},
        xtick=data]
        \addplot[mark=diamond*,thick,red] coordinates {
           (1, 0.840)    (2, 0.842)    (32, 0.854)    (64, 0.855)    (128, 0.859)    (256, 0.860) (384, 0.859) (511, 0.860)};
        \addlegendentry{NetVLAD}
        \addplot[mark=o,mark options={solid},blue,thick,dashed] coordinates {
           (1,0.840)    (2, 0.844)    (32,0.858)    (64, 0.860)    (128, 0.863)    (256, 0.864) (384, 0.864) (511, 0.863)};
        \addlegendentry{DSGMM}
    \end{axis}
    \end{tikzpicture}
    
    \end{center}
    \caption{Impact of number of clusters on model performance - GAP and Hit@1}
    \label{fig:varyk}
\end{figure*}
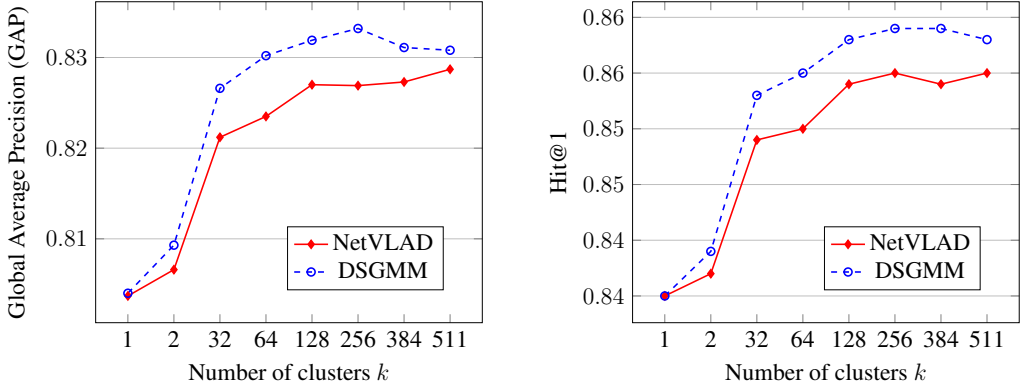

Figure \ref{fig:varyk} depicts the relationship between GAP/Hit@1 and the number of clusters for both NetVLAD and DSGMM. We train these models with $\gamma = 1.0$ and without any normalization. We note that DSGMM outperforms NetVLAD across all settings of $k$. Both models significantly improve in performance as $k$ is increased from 1 to a larger value, although performance starts to saturate at $k=256$ for both models. The maximum value of $k$ we were able to successfully use is 511, constrained by practical considerations like the model size on the filesystem. We also note that DSGMM with 64 clusters outperforms NetVLAD with 256 clusters. This suggests that DSGMM could help learn more compact video representations without sacrificing performance. 

\textbf{Effect of normalization}. Given a representation of size $K \times D$, we consider two different kinds of normalizations. We first consider L2-normalizing each row of the representation \cite{allaboutvlad}, known as \textit{intra-norm}. We also consider L2-normalizing the entire video representation. We refer to this normalization as  \textit{final-norm}. 

We report results for NetVLAD and DSGMM architectures across all possible combinations of the aforementioned normalization techniques in Table \ref{tab:varynorm}, with $K=256$. We find that DSGMM outperforms NetVLAD in all settings except one.

\textit{Effect of intra-norm.} We find that \textit{intra-norm} helps improve performance across the board, with or without \textit{final-norm}.
 
\textit{Effect of final-norm.} The effect of the \textit{final-norm} is not as clear. While it hurts DSGMM and helps NetVLAD for the setting without \textit{intra-norm}, it seems to have minimal effect on \textit{intra-normed} embeddings for both NetVLAD and DSGMM.

\begin{table}
\begin{center}
\scalebox{0.85} {
\begin{tabular}{llcccc}
\toprule
\textbf{I-N} & \textbf{F-N} & \multicolumn{2}{c}{\textbf{GAP}} & \multicolumn{2}{c}{\textbf{Hit@1}} \\
\cmidrule{3-6}
 & & NetVLAD & DSGMM & NetVLAD & DSGMM \\
\midrule
No & No & 0.826 & \textbf{0.833} & 0.859 & \textbf{0.864} \\ 
\midrule
Yes & No & 0.831 & \textbf{0.834} & 0.863 & \textbf{0.864} \\ 
\midrule
No & Yes & \textbf{0.830} & 0.829 & 0.861 & 0.861 \\ 
\midrule
Yes & Yes & 0.831 & \textbf{0.834} & 0.863 & \textbf{0.865} \\
  \bottomrule
  \end{tabular}
  }
  \vspace{0.5mm}
    \caption{Effect of intra and final norm on model performance. I-N is used to designate \textit{intra-norm}, F-N for \textit{final-norm}.}
    \label{tab:varynorm}
\end{center}
\end{table}

\subsubsection{DSGMM with varying $\gamma$} \label{varygamma}
\begin{figure*}
\begin{center}
\begin{tikzpicture}[scale=0.75][font=\large]
\begin{axis}[
    legend style={
            at={(0.45, 0.25),anchor=north}
        },
   xlabel={$\log_2\gamma$},
    ylabel={Global Average Precision (GAP)},
     ymajorgrids = true,
    symbolic x coords={-9, -6, -3, -2, -1, 0, 1, 2, 3, 4, 5, 7},
    xtick=data]
    \addplot[mark=o,thick,purple] coordinates {
        (-9,0.8318) (-6,0.8334) (-3,0.8353) (-2,0.8343) (-1,0.8347) (0,0.8345) (1,0.8291) (2,0.8235) (3,0.8176) (4,0.8155) (5,0.8131) (7,0.7944) };
    \addlegendentry{DSGMM}
    \addplot[mark options={solid},blue,thick,dashed] coordinates {
        (-9,0.8307) (7, 0.8307) };
    \addlegendentry{NetVLAD}
\end{axis}
\end{tikzpicture}
\qquad
\begin{tikzpicture}[scale=0.75][font=\large]

\begin{axis}[
    legend style={
            at={(0.45, 0.25),anchor=north}
        },
    xlabel={$\log_2\gamma$},
     ymajorgrids = true,
    ylabel={Hit@1},
    symbolic x coords={-9, -6, -3, -2, -1, 0, 1, 2, 3, 4, 5, 7},
    xtick=data]
    \addplot[mark=o,thick,orange] coordinates {
        (-9,0.863) (-6,0.864) (-3,0.866) (-2,0.865) (-1,0.865) (0,0.864) (1,0.860) (2,0.856) (3,0.853) (4,0.854) (5,0.854) (7,0.841) };
    \addlegendentry{DSGMM}
    \addplot[mark options={solid},blue,thick,dashed] coordinates {
        (-9,0.863) (7,0.863) };
    \addlegendentry{NetVLAD}

\end{axis}
\end{tikzpicture}

\end{center}
    \caption{Impact of $\gamma$ on DSGMM performance - GAP and Hit@1}
    \label{fig:varygamma}
\end{figure*}
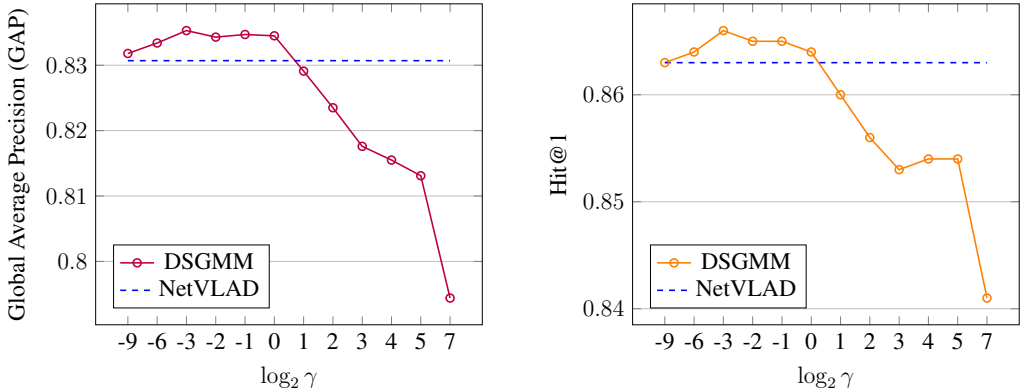

As Equation \ref{eq:lambda-formulation} shows, $\gamma \geq 0$ is a smoothing hyperparameter between the UBM and the mean sufficient statistic for an individual video. In Figure \ref{fig:varygamma}, we note the empirical performance of DSGMM models with increasing $\gamma$. Normalization is set such that we are doing \textit{intra-norm} but not \textit{final-norm}. $K=256$ in these experiments. On the horizontal axis, we traverse through different powers of two that $\gamma$ can take in order to widen the search space. As shown, for higher values of $\gamma$, performance does degrade. Similarly as $\gamma$ decreases to zero, there is a drop but not as severe. At exactly $\gamma=2^{-\infty} = 0$ which is not shown on the plot, we observe a performance of $83.0\%$ GAP and $86.1\%$ Hit@1, which is comparable to the $\gamma=2^{-9}$ datapoint plotted. The horizontal lines in both plots show NetVLAD performance for reference. $\gamma=2^{-3}=0.125$ seems to be the best performant smoothing parameter in these experiments although the difference between $\gamma = 2^{-3}, 2^{-2}, 2^{-1}$ is small.

\subsubsection{NetVLAD and DSGMM performance across GMM model type}
\begin{table}
\begin{center}
\scalebox{0.85} {
\begin{tabular}{lccccc}
\toprule
\textbf{Model Type} & \multicolumn{2}{c}{\textbf{GAP}} & \multicolumn{2}{c}{\textbf{Hit@1}} \\
\cmidrule{2-5}
 & NetVLAD & DSGMM & NetVLAD & DSGMM \\
\midrule
\texttt{Decoupled} & 0.831 & \textbf{0.835} & 0.863 & \textbf{0.865} \\ 
\midrule
\texttt{UniformPriors} & 0.831 & \textbf{0.832} & 0.860 & \textbf{0.862} \\ 
\midrule
\texttt{SharedSpherical} & 0.830 & \textbf{0.834} & 0.860 & \textbf{0.864} \\  
\midrule
\texttt{Spherical} & 0.830 & 0.830 & \textbf{0.861} & 0.860 \\ 
\midrule
\texttt{SharedDiagonal} & 0.830 & \textbf{0.833} & 0.860 & \textbf{0.864} \\  
\midrule
\texttt{Diagonal} & 0.830 & \textbf{0.832} & 0.859 & \textbf{0.861} \\  
  \bottomrule
  \end{tabular}
  }
  \vspace{0.5mm}
    \caption{NetVLAD and DSGMM performance for different GMM model types - GAP and Hit@1.}
    \label{tab:gmmtype}
  \vspace{-6mm}
\end{center}
\end{table}

Based on the GMM frameworks described in Section \ref{subsec:NetVLAD}, we run a series of experiments training two models for each type: NetVLAD and DSGMM. Across experiments, the setting is kept the same: $K = 256$, $\gamma=0.5$, with \textit{intra-norm} but no \textit{final-norm}. The number of parameters do differ across the different models. Focusing on the clustering component, the \texttt{Decoupled} approach has $2KD + K$ parameters. This is only matched by the \texttt{Diagonal} approach. From Table \ref{tab:gmmtype}, we observe that in the majority of cases, DSGMM outperforms NetVLAD in terms of GAP and Hit@1. The \texttt{Decoupled} approach performs the best out of the model types for NetVLAD and DSGMM.

\subsection{Results on video recommendation using similarity-based aggregation} \label{sec:resultssocialnetwork}

We compare different video representations on the task of video recommendation for LinkedIn users. We use user-video watch data over a period of 4 weeks to construct training and testing datasets. We collect binary labels $\{0, 1\}$ on whether a user $U$ will watch a video $v_c$ above a threshold of $T$ seconds or not. We generate scalar similarity scores between a user and candidate video using equations \ref{eq:avgsim} and \ref{eq:maxsim}, and then compute area under ROC curve (AUC) using these scores and the labels. 

 Apart from NetVLAD and DSGMM, we consider average pooling of frames as a baseline representation. From Table \ref{tab:socialnetworkdata}, it is evident that a 2-layer neural network used in conjunction with the DSGMM architecture outperforms NetVLAD.  
 
 In this section and Section~\ref{sec:glmixexperiment}, we only report relative percent improvements because the raw numbers are sensitive company data.

\begin{table}
  \begin{center}
\scalebox{0.85} {
\begin{tabular}{lcc}
\toprule
\textbf{Model} & \textbf{Max sim.} & \textbf{Avg sim.} \\
\midrule
Avg Pooling of Frames & - & - \\
\midrule
2-layer NN + NetVLAD & +16.96\% & +15.69\% \\
\midrule
2-layer NN + DSGMM & \bf{+17.84\%} & \bf{+19.26\%} \\
  \bottomrule
  \end{tabular}
  }
  \vspace{0.5mm}
  \caption{Relative AUC improvement from Avg Pooling of Frames for predicting user-video affinity. We consider both average and maximum similarity aggregation over a user's watch history.}
    \label{tab:socialnetworkdata}
    
    \vspace{8mm}
  
\scalebox{0.85} {
\begin{tabular}{lcc}
\toprule
\textbf{Video Feature} & \textbf{All} & \textbf{Cold-start} \\
\midrule
No video feature & - & -\\
\midrule
2-layer NN  + NetVLAD & +3.69\%  & +4.98\%\\
\midrule
2-layer NN  + DSGMM & \bf{+3.85\%}  & \bf{+5.44\%}\\
  \bottomrule
  \end{tabular}
  }
  \vspace{0.5mm}
  \caption{Relative AUC improvement from a baseline GLMix model without video features for predicting video watches.}
    \label{tab:glmix}
\vspace{-8mm}
\end{center}
\end{table}

\subsection{Results on video recommendation with the GLMix model}\label{sec:glmixexperiment}
The GLMix model with different video features is also evaluated in Table~\ref{tab:glmix}, i.e. the same task as in Section~\ref{sec:resultssocialnetwork}. We also focus on cold-start videos that are fresh to the video recommendation system. For these types of videos, the video feature is more important in the recommendation task because of unavailability of features like video popularity. As shown in Table~\ref{tab:glmix}, video features help improve performance, especially in the cold-start case. Furthermore, the architecture with the $2$-layer neural network used in conjunction with the DSGMM architecture outperforms NetVLAD.

\section{Conclusion and Future Work} \label{sec:conclusion}

In this paper, we address the problem of video representation learning using Gaussian mixture models, developing two techniques called SGMM and DSGMM. Our methodology relies on smoothing the representation for a video in a cluster when the cluster receives little or no data from the video. We demonstrate the efficacy of our methods on the YouTube-8M classification dataset~\cite{youtube8m}, and predicting whether a LinkedIn user will watch a video presented on their feed. In all cases, our new approach was superior to NetVLAD.

For future work, we would like to include audio features in our methodology to further improve the performance. We would also like to test the efficacy of the proposed DSGMM approach on other video related tasks like action recognition, and video content summarization. 


{\small
\bibliographystyle{apalike}
\bibliography{egbib}
}

\end{document}